\documentclass[square]{article}

\PassOptionsToPackage{numbers, compress}{natbib}
\usepackage[preprint]{neurips_2023}
\usepackage{mdframed}
\usepackage{graphicx}
\usepackage{color,soul}
\usepackage{tabularx,multicol,multirow}
\usepackage[most]{tcolorbox}
\usepackage{doi}

\usepackage[utf8]{inputenc} 
\usepackage[T1]{fontenc}    
\usepackage{hyperref}       
\usepackage{url}            
\usepackage{booktabs}       
\usepackage{amsfonts}       
\usepackage{nicefrac}       
\usepackage{microtype}      
\usepackage{xcolor}         

\title{Designing Heterogeneous LLM Agents for \\Financial Sentiment Analysis}

\author{%
  Frank Xing \\
  School of Computing\\
  National University of Singapore\\
  \texttt{xing@nus.edu.sg} \\
}

\begin{document}

\maketitle

\begin{abstract}
Large language models (LLMs) have drastically changed the possible ways to design intelligent systems, shifting the focuses from massive data acquisition and new modeling training to human alignment and strategical elicitation of the full potential of existing pre-trained models. This paradigm shift, however, is not fully realized in financial sentiment analysis (FSA), due to the discriminative nature of this task and a lack of prescriptive knowledge of how to leverage generative models in such a context. This study investigates the effectiveness of the new paradigm, i.e., using LLMs without fine-tuning for FSA. Rooted in Minsky's theory of mind and emotions, a design framework with heterogeneous LLM agents is proposed. The framework instantiates specialized agents using prior domain knowledge of the types of FSA errors and reasons on the aggregated agent discussions. Comprehensive evaluation on FSA datasets show that the framework yields better accuracies, especially when the discussions are substantial. This study contributes to the design foundations and paves new avenues for LLMs-based FSA. Implications on business and management are also discussed. 
\end{abstract}

\section{Introduction}
Since OpenAI's ChatGPT went viral one year ago, large language models (LLMs) have gone through fast improvements, showing a variety of capabilities. The AI adaptation for many financial services is accelerating, and big data-supported financial decision-making is no exception. Financial sentiment analysis (FSA) is a prototypical task in that category and is becoming increasingly important as financial service processes and our social behavior digitalize: companies disclose electronic versions of their annual reports, earning calls, and announcements, and investors join online communities, discussion forums, and social media to interact with others. The recent GameStop Saga~\cite{Deng2023} and the popularity of a spectrum of market sentiment indexes (e.g., MarketPsych~\cite{Peterson2016}) have shown clear evidence that sentiment is a useful analytics tool for financial decision-making, forecasting short-term returns and volatilities~\cite{saving}, detecting fake news and fraud~\cite{Dong2018}, and predicting risk~\cite{yang2023}. The usefulness and the importance of accurate FSA are also underpinned by a long thread of research~\cite{Bollen2011,xing2018,Deng2018,chu2022}. Hendershott et al.~\cite{h2021} summarized that research on the application of AI on news, social media, and word-of-mouth data is a major category of leveraging AI in finance. Considering these factors, accurate FSA is desired for multiple stakeholders. 

The majority of FSA systems were developed in the past decade and their architecture and design ideas have gone through several iterations along with the advances in natural language processing. Early systems rely on sentiment word dictionaries and simple rules or statistics to derive sentence-level or message-level polarities. Efforts were made to discover words/phrases specific to the finance domain~\cite{loughran2011,xing2019cognitive}. A great amount of learning-based systems were later developed. Specifically, two benchmark tasks (SemEval 2017 Task 5~\cite{cortis2017semeval} and FiQA 2018 Task 1~\cite{de2018inf}) were conducted, and the best results were achieved by regression ensemble (RE), convolutional neural network (CNN), and support vector regression (SVR) models based on combined features of sentiment lexica and dense word representations. The following wave of designs were based on fine-tuning general-purpose pre-trained language models, e.g., BERT\footnote{There is no strict definition of ``how large" a language model has to be to qualify for the name of LLM. It seems that LLMs are usually far larger than the word2vec models (around 1 million parameters). This definition includes BERT (110-340 M parameters), GPT-3 (around 175 B parameters), and more.} (Bidirectional Encoder Representations from Transformers). For example, FinBERT~\cite{Liu2020} achieved good FSA results and the state-of-the-art is from integrating multiple auxiliary knowledge sources to a BERT variant~\cite{duk2023}. In terms of leveraging LLMs for FSA, the current progress mainly employed the encoder type of transformer, e.g., BERT. However, the most powerful LLMs now are based on the decoder part of a transformer. The decoder architecture is natural for generative tasks such as discourse/chat completion and question answering, but can also be fitted for discriminative tasks and classification. This study is aware of the early stage and scant in-depth studies in this direction and thus explores ways of leveraging generative LLMs for FSA. 

Different from many ad hoc designs developed from chain of thought (CoT)~\cite{deng-www}, tree of thoughts (ToT)~\cite{tot}, verification, self-consistency constraints, intermediate scratchpads, and multi-agent multi-role settings, the design framework presented here follows the design science guidelines by Hevner et al.~\cite{hevner} and contributes to the prescriptive knowledge as a ``design theory"~\cite{gregor}. Based on Minsky’s theory of mind and emotions, ``emotional states" are our Ways to Think with a specific collection of resources turned on and others turned off given certain environment conditions~\cite{emotion-machine}. Therefore, one FSA approach is to simulate the mental processes underlying the texts, requiring specialized LLM agents to play the roles of ``resources", i.e., functional parts of our brain that make us react to the environment. In the context of financial analysis, many resources are learned as professional knowledge and not innate parts of our brains. The design framework (\textbf{H}eterogeneous multi-\textbf{A}gent \textbf{D}iscussion) chooses to develop specialized LLM agents by prompting, and the main function is to pay attention to a type of error that LLMs are prone to make for the given FSA task. The design artifact thus has five different agents and the FSA result is based on a shared discussion considering output from all the agents. I evaluate the artifact using multiple methods, and the results generally conclude the framework to be effective.  

\begin{figure}[ht]
\centering
\includegraphics[width=1.05\textwidth]{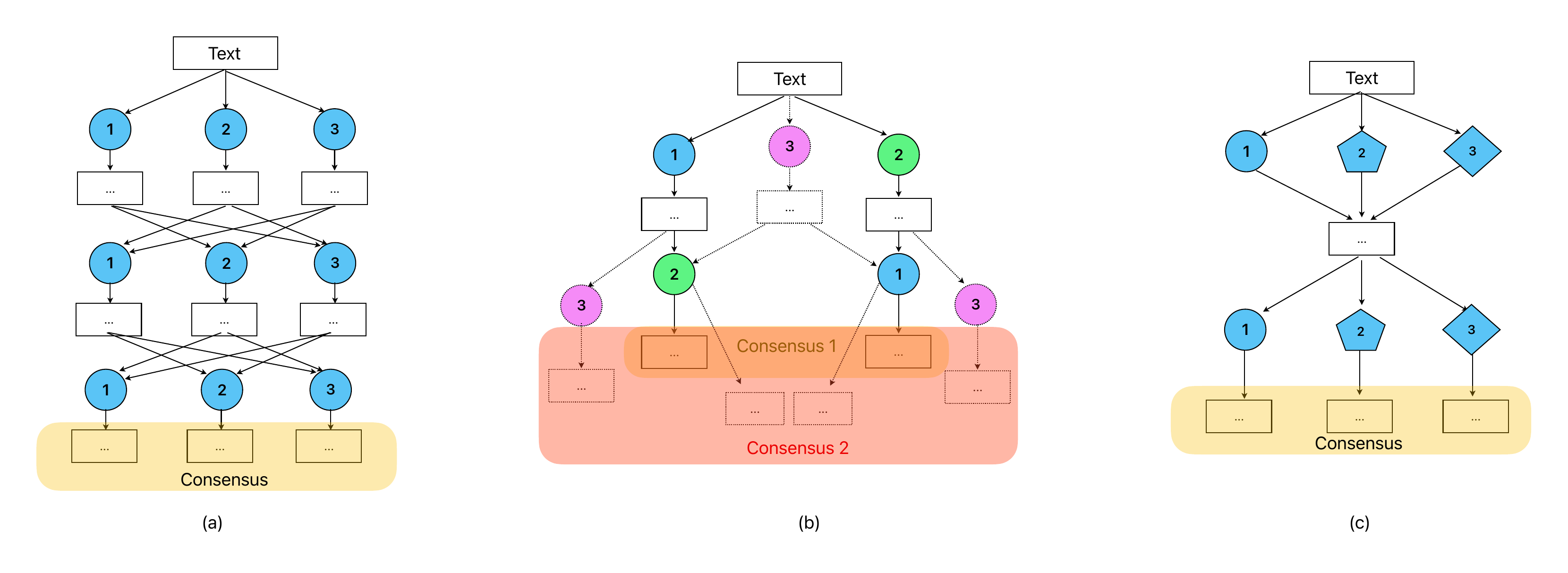}
\caption{Different multi-agent LLM frameworks for reaching a consensus: (a) homogenous multi-agent debate~\cite{md}, (b) multi-role multi-agent negotiation~\cite{mn}, (c) heterogeneous multi-agent discussion (HAD: the proposed framework). Colors denote different roles and shapes denote heterogeneous agents.}
\label{fig-comp}
\end{figure}

The major challenge in instantiating this design is the lack of design theory on what each agent's function should be. For this reason, many LLM multi-agent settings employ homogeneous agents. For example in the multi-agent debate framework, Du et al.~\cite{md} simply disseminate the same input to multiple LLM agents. Because of some randomness and perturbation, each agent's response will not be identical. Later each agent will take outputs from other agents (excluding its own output) as additional information to update its original response (Fig.~\ref{fig-comp} (a)). It may go through multiple rounds though empirical results show that consensus will be achieved fast. Another framework is to assign different roles to LLM agents. Sun et al.~\cite{mn} described a negotiation procedure where a ``discriminator LLM" is asked to judge whether it agrees with the output of a ``generator LLM". The judgment statement is sent back to the generator if consensus is not made. The framework requires a third LLM to negotiate and vote for the final result if discrepancies persist (Fig.~\ref{fig-comp} (b)). Although the LLM agents in this framework play different roles, their capability assumptions remain the same. In such a sense, these agents are still non-specialized and homogeneous. In the proposed framework (Fig.~\ref{fig-comp} (c)), each agent has the same role, goes through a symmetric discussion workflow (unlike~\cite{mn}), but is purposely designed to simulate the mental functions of different resources. Their responses are aggregated for FSA just like resources are activated to generate different emotional states.  

Therefore, one objective of this study is to test whether error types~\cite{Zimbra,fsa} can be a useful guideline for developing heterogeneous agents. Specifically, I am interested in the following research questions:
\begin{itemize}
    \item RQ1: How effective is HAD compared to naive prompting and to the fine-tuning paradigm?
    \item RQ2: How to prompt LLM agents to behave heterogeneously for sentiment analysis in finance?
    \item RQ3: What are the quantitative contributions of each LLM agent and their relative importance?
\end{itemize}

To address these questions, HAD is evaluated using multiple methods including empirical analysis of performance metrics on five FSA datasets, ablation analysis with different sets of agents, and case studies of outputs and intermediary representations. The experimental results show that HAD can in general improve the FSA performance and the improvements are constant for GPT-based LLM agents. It has been observed that a simple template ``please pay special attention to [error type]" can change LLM agents' attention and prompt them to behave differently. Mood, rhetoric, and reference agents seem to be the main performance drivers and are more critical than other LLM agents, though the contributions are non-linear and have complicated interactions. 

This study contributes to the design science literature by presenting a kernel theory-informed design artifact. A number of kernel theories from the natural or social sciences were introduced to information system design, whereas kernel theories from AI are comparatively rare. This study has implications for the emotion theory, LLM collaboration research, and financial decision-making practices. Firstly, it supports the society of mind and emotion machines~\cite{emotion-machine} to be actionable theories that explain how emotions emerge as an important type of human intelligence; Secondly, this study applies multi-agent LLMs in FSA. This framework has been used for factuality checking, arithmetic/mathematical reasoning, optimization, general-purpose sentiment analysis, but not yet on FSA to the best of my knowledge. This study thus provides new materials for LLM collaboration, and also reinforces the design science-based approach to framework development; Lastly, the findings contribute to the prescriptive knowledge of FSA system design. Investors and traders may iterate and improve their own FSA systems based on the HAD framework or be more informed when they decide to select or purchase technical solutions of a similar kind. 


\section{Related Work and Design Process}
In this section, related literature is organized into two lines: (any type of) use of LLMs for FSA, and ways of prompt design (not limited to FSA). I also elaborate on the theoretical foundations of employing heterogeneous agents for FSA. 

\subsection{Using LLMs for Financial Sentiment Analysis}
Financial sentiment analysis (FSA) is a domain-specific business-oriented application closely related to the general natural language processing task of sentiment analysis. Because of its heavy use of terminologies and other linguistic features~\cite{fsa,shah2022}, general sentiment analysis performances are usually not representative and will drop in the finance domain. FSA has been included to comprehensively evaluate LLMs for finance, together with other tasks such as Name Entity Recognition (NER), knowledge recall, question answering, and reading comprehension among others~\cite{shah2022,bloomberggpt}. 

In terms of using a singular LLM, the FSA task is sometimes formulated together with target or aspect detection, and the additional information may be used to improve FSA performances. For example, Lengkeek et al.~\cite{lengkeek} used the hierarchical structure of aspect systems to constrain FSA results, though this information is rarely available in real-world production environments. Zhang et al.~\cite{zhang-icaif} observed that financial news is often overly succinct. A model that retrieves additional context from reliable external sources to form a more detailed instruction is consequently developed. Deng et al.~\cite{deng-www} found that forcing the LLM through several reasoning paths with CoT helps generate more stable and accurate labels. The LLM generated labels are also useful and meet the quality for complementing human annotations for conventional supervised learning methods. Similarly, Fei et al.~\cite{fei2023} developed a three-hop reasoning framework inspired by CoT that infers firstly the implicit aspect, secondly the implicit opinion, and finally the sentiment polarity. However, it has been pointed out~\cite{mn} that a singular LLM has difficulties in fully exploiting the potential of LLM knowledge. This is especially true for FSA as it involves multiple LLM capabilities, such as reasoning, fact-checking, syntactic/semantic parsing, and more. I observe a similar phenomenon as reported in~\cite{zhangd} that LLM performances on more complicated tasks are not as satisfactory as on the binary classification task. Moreover, the aforementioned designs (storage retrieval and CoT) and more designs that are not yet applied to FSA, such as verification, self-consistency constraints, or intermediate scratchpads, are also largely heuristic, at most based on experiences, and lack solid theoretical foundation.

The proposed framework adopts in-context learning (ICL) and leverages multiple LLM instantiations (agents), which is also referred to as LLM collaboration. Strategies of collaboration include auxiliary tasks (e.g., verification)~\cite{chen23b}, debate~\cite{md}, and various role-assignment~\cite{mn} including generator, discriminator, programmer, manager, meta-controller, etc. Again, the design of auxiliary tasks and roles appears arbitrary and lacks solid theoretical foundations. LLM collaboration is also more investigated on many general natural language processing tasks including sentiment analysis, but their applicability on FSA lacks direct evidence. Perhaps most related to the proposed HAD design framework is MedPrompt~\cite{nori23}. It uses an ensemble of randomly shuffled CoT from homogeneous agents. The design is also more computationally heavy and difficult to transfer to the finance domain as existing financial question-answering datasets are more sparse.  

\subsection{Prompt Engineering}
Before the emergence of generative LLMs, a well-accepted way of applying a language model to downstream tasks is through fine-tuning: remove the last neural network layer (referred to as the ``head" layer) and let the training errors back-propagate with the bottom layers parameters fixed. Two major problems with it are: (1) a not-too-small training set and labels are still needed, and (2) the training process can be computationally intensive. With the observation that generative LLMs are very powerful, in-context learning contends it possible to get the desired output without fine-tuning and elicit the model capability with an appropriate ``prompt". Typically, prompt engineering involves the development of task-specific prompt templates, which describe how a prompt should be formulated to enable the pre-trained model to perform the downstream task at hand. Liu et al.~\cite{liupf} provided a survey on recent advances in prompt engineering and systematically compared cloze prompts and prefix prompts.

In terms of automatically searching for the prompt template, stochastic optimization-based methods are discussed. Sorensen et al.~\cite{Sorensen2022}, for example, discovered that a good template is the one that maximizes the mutual information between input and the generated output. 

In terms of designing prompts, Liu and Chilton~\cite{Liu2022} studied text-to-image generative models and the prompt template ``SUBJECT in the style
of STYLE". They found the clarity and salience of keywords are important to the generation quality. Yu et al.~\cite{yu2023} presented the idea of using domain knowledge to guide prompt design. It was reported that for the legal information entailment task, the best results are obtained when prompts are derived from specific legal reasoning techniques, such as Issue-Rule-Application-Conclusion (IRAC) as taught at law schools. For FSA, however, the design guidelines are unclear and most studies used naive prompts. For example, Chen and Xing~\cite{chen23} used ``You are a helpful sentiment analysis assistant - [example message]:[sentiment]. User: [test message]." and BloombergGPT's FSA template~\cite{bloomberggpt} is simply ``[test message] Question: what is the sentiment? Answer with negative/neutral/positive". 

\subsection{Kernel Theory: Emotions and the Society of Mind}
Kernel theory is a key component of the information system design process according to Walls' information system design theory (ISDT). It explains how/why the anticipated system would work and sheds light on the meta-requirements. In the context of FSA, the theory has to be one that explains the formative mechanism of emotion. For this reason, Minsky's theory of mind and emotions is preferred over other descriptive/contrastive theories of emotions, such as Plutchik's wheel of emotions or Russell's circumplex model.

\begin{figure*}[t]
\centering
\includegraphics[scale=1.2]{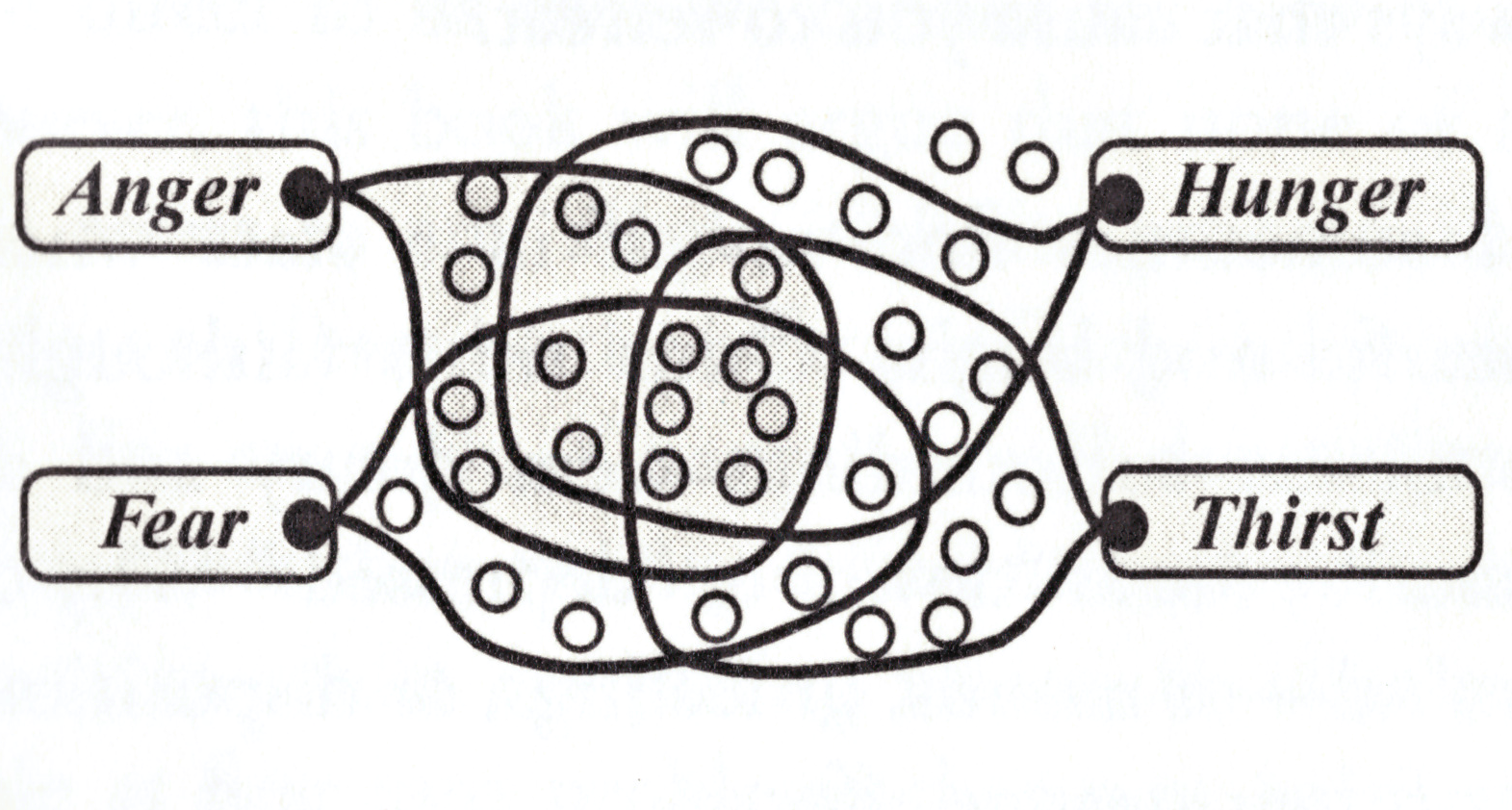}
\caption{Illustration of the generation of emotional states from activating a collection of resources, cf. pg. 4 in~\cite{emotion-machine}.}
\label{fig:emo}
\end{figure*}

Society of mind is a reductionistic perspective of human intelligence that influenced AI greatly and argues no function directly produces intelligence. Instead, intelligence comes from the managed interaction of a variety of resourceful but simpler and non-intelligent agents. For example, when drinking a cup of tea, there activates a motor agent that grasps the cup, a balancer that keeps the tea from spreading, and a temperature sensor that confirms our throat will not be hurt. This theory sees emotional states as patterns of activation. For example, the state we call ``angry" could be what happens when a cloud of resources that help you react with unusual speed and strength are activated --- while some other resources that make you act prudently are suppressed (Fig.~\ref{fig:emo}).

\begin{table}[b]
\centering
\caption{Kernel Theory-Based Design: A Meta-Framework}
\label{tab:table}
\resizebox{\columnwidth}{!}{%
\begin{tabular}{|c|l|}
\hline
\textbf{Kernel theory} & Society of mind and emotion machines.\\& The theories posit that emotions come from activation of different resources.\\ \hline

\textbf{Meta-requirements} &
  \begin{tabular}[c]{@{}l@{}}1. To simulate the resources, we should define agents and their functions. \\ 2. To activate the agents, we should provide information about the subjectivity.
  \\ 3. To achieve a well-informed decision, we should aggregate information from different agents.
  \end{tabular} \\ \hline

\textbf{Meta-designs} &
  \begin{tabular}[c]{@{}l@{}}1. Types of error are used as domain knowledge to guide building heterogeneous agent capabilities. \\ 2. The user message is distributed to each LLM agent.\\ 3. Specialized agent outputs are concatenated to form the summative prompt.
  \end{tabular} \\ \hline

\textbf{Testable hypotheses} &
  \begin{tabular}[c]{@{}l@{}}Evaluate the effectiveness of the metadesigns. Specific testable hypotheses are as follows:\\ Hypothesis 1: The HAD framework can improve the accuracy of
existing naive prompts for FSA.\\ Hypothesis 2: The agents have different importance but all contribute positively to the analysis.
\end{tabular} \\ \hline
\end{tabular}%
}
\end{table}

Minsky's theory of emotion posits that you feel ``angry" when your cake is stolen by other kids, because the IF-THEN-DO rules activate resources to help you take it back. The activation is adaptive as we learn and grow. For FSA, a crucial procedure is to decide what candidate resources need to be designed: it will not require the full set of resources in our brain which will be more challenging to build. In the remainder of this section, I describe the design rationales using a kernel theory-based design science framework (Table~\ref{tab:table}).

\subsection{Meta-requirements, Meta-designs, and Hypotheses}
\begin{figure}[htp]
\centering
\includegraphics[scale=1.05]{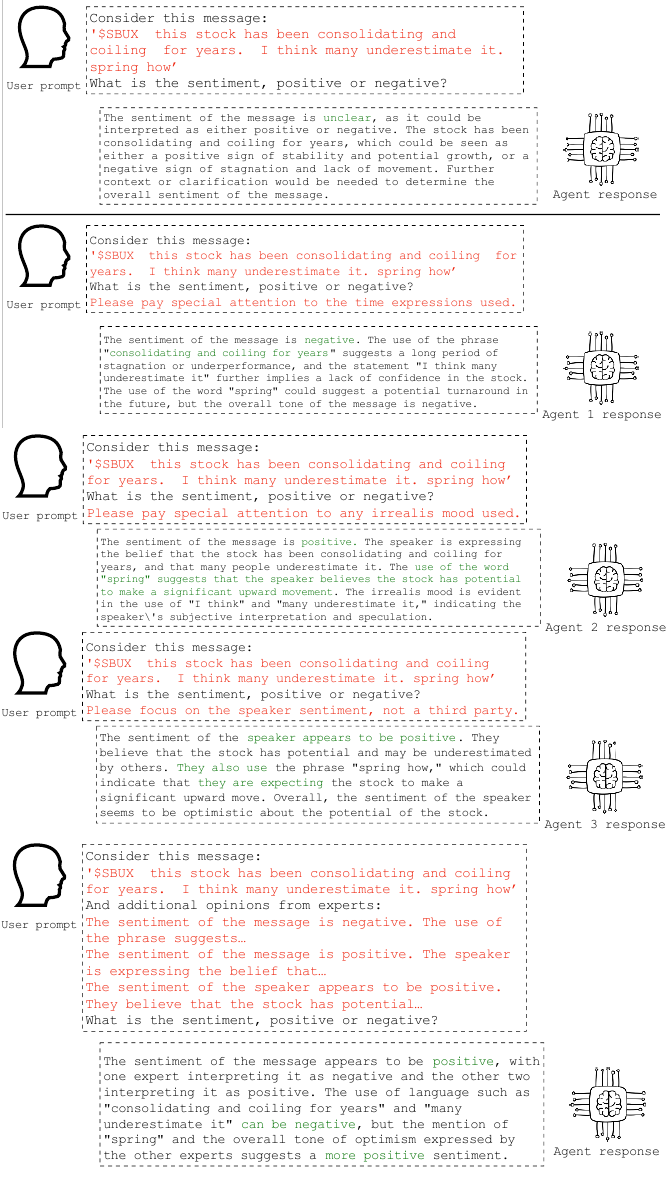}
\caption{An illustrative comparison between naive prompting (the upper example) and the proposed HAD framework (the lower example) with 3 heterogeneous agents inspired by FSA error types.}
\label{fig:illu}
\end{figure}

Although the society of mind relies heavily on the conceptual construct of ``resource", it is purposefully kept in a hazy way (pg. 25 in~\cite{emotion-machine}), referring to all sorts of functional parts that range from perception and action to reflective thinking. Therefore, it seems appropriate to simulate the resources using LLM agents with polymathic capabilities, and specialize their functions via prompts.
This choice also makes resource activation plausible, because specialized agents will not generate meaningful responses to the out-of-scope context. It is thus designed such that all the LLM agents will receive the original user message. To aggregate information, a widely used technique is to concatenate them into a longer prompt~\cite{mn,md,Hendrycks,liupf}. By translating the meta-requirements into more detailed meta-designs, the HAD framework can be formally represented as:
\begin{enumerate}
    \item Define heterogeneous agents and their prompt templates $A_1$, $A_2$, ..., $A_k$.
    \item Obtain intermediary analysis $O_i = A_i (User\_Message)$
    \item Obtain summative analysis $Result = A (User\_Message, O_1,...,O_k)$
\end{enumerate}
The second step can be carried out for multi-rounds before sending for a summary depending on the consensus situation. An illustration of the workflow is presented in Fig.~\ref{fig:illu}.

To assess whether the proposed framework is effective, two testable hypotheses are developed. If types of error are useful for guiding agent design, we would expect the performance metrics to improve (\textbf{H1}). Because of the noted data imbalance issue in FSA, F-1 score should also be investigated on top of accuracy. Another observation is that the occurrences of each type of error are not equal and vary across different language domains~\cite{Zimbra,fsa}. It is thus hypothesized that the agents will have different importance but all contribute positively to the FSA task (\textbf{H2}).

\section{Design Artifact: Heterogeneous Agent Discussion (HAD)}
To instantiate a design artifact based on the HAD framework, the number of agents ($k$) has to be decided. 

Zimbra et al.~\cite{Zimbra} had investigated a comprehensive list of Twitter sentiment analysis methods and concluded three major challenges: (1) language brevity, (2) imbalanced classes, and (3) temporal dependency. Because of these challenges, 13 categories of commonly occurring classification errors were identified. The main categories that ground to linguistic features can be summarized as: (1) humor, (2) subtlety or a mixture of sentiment, (3) irrelevance (e.g., aspect mismatch), (4) marketing information mistaken as positive, and (5) atypical contextual usage. Xing et al.~\cite{fsa} investigated the common errors in a slightly different scope: specifically for FSA and including text sources other than Twitter\footnote{Now has been re-branded as ``X".}. The 6 categories of errors identified, i.e., (1) irrealis mood, (2) rhetoric, (3) dependent opinion, (4) unspecified aspects, (5) unrecognized words, and (6) external reference, have significant overlap with those reported from~\cite{Zimbra}. 

With this background, five agents are designed based on~\cite{fsa} because (1) these categories are more directly FSA relevant and (2) these categories are less in number (6 compared to 13) and more operational. Since LLMs are observed to be robust to unrecognized words and spellings from the web, no special agent is designed according to this error. The five agents and their characteristic prompts are:
\begin{itemize}
    \item \textbf{A1} (the mood agent): Please pay special attention to any irrealis mood used.
    \item \textbf{A2} (the rhetoric agent): Please pay special attention to any rhetorics (sarcasm, negative assertion, etc.) used.
    \item \textbf{A3} (the dependency agent): Please focus on the speaker sentiment, not a third party.
    \item \textbf{A4} (the aspect agent): Please focus on the stock ticker/tag/topic, not other entities.
    \item \textbf{A5} (the reference agent): Please pay special attention to the time expressions, prices, and other unsaid facts.
\end{itemize}
The summative prompt takes the form of ``Considering this message: [test message] and additional opinions from experts [opinions], what is the sentiment, positive/negative/neutral?". Some nuances are adjusted according to whether the testbed classification is binary or ternary.

\section{Evaluation}
Hevner et al.~\cite{hevner} described five kinds of design evaluation methods. This study leverages three out of them: (1) empirical testing on existing datasets and the produced performance metrics, (2) ablation analysis with manipulated module components, and (3) observational evaluation based on case studies.

\subsection{Performance Improvement}
The proposed design framework is evaluated on five existing datasets, i.e., the Financial PhraseBank~\cite{malo2014good}, StockSen~\cite{fsa}, CMC~\cite{chen23}, FiQA Task 1~\cite{maia201818}, and SEntFiN 1.0~\cite{sentfin}. The last two are finer-grained financial sentiment analysis datasets with sentiment intensity scores or multiple targets/entities labels, though I applied quantization and filtering to fit the evaluations into a consistent classification problem. For example, the original FiQA dataset~\cite{maia201818} has 1173 messages with sentiment scores ranging from -1 to +1. By filtering those scores with an absolute value larger than 0.3, only 771 messages are left and mapped to the positive/negative classes. The detailed statistics are reported in Table~\ref{stats}. In terms of text genre, Financial PhraseBank (FPB) is from news and SEntFiN is from news headlines. StockSen and CMC are from social media (StockTwits and CoinMarketCap.com respectively), and the whole FiQA is consolidated from crawling a mix of StackExchange, Reddit, and StockTwits.

The HAD framework is tested on two instruction-finetuned language models: GPT-3.5\footnote{https://platform.openai.com/docs/models/gpt-3-5} as a restrict-access representative, and BLOOMZ\footnote{https://huggingface.co/bigscience/bloomz} (the 560m version~\cite{bloomz}) as an open-access representative. The performance metrics on GPT-3.5 are obtained through OpenAI API, and metrics on BLOOMZ are computed using a laptop with an 8-core Apple M1 chip and 16 GB memory. For both LLMs, one experiment takes minutes to hours to execute. The performance metrics are reported in Table~\ref{tab:freq}. For ternary classifications (FPB and SEntFiN), macro F-1 scores are used. Some metrics (in grey color) of BloombergGPT~\cite{bloomberggpt} and (Fin-)BERT~\cite{fsa,shah2022,duk2023,chen23,sentfin} are included to help roughly assess the gaps to fine-tuning based results. Noteworthy, these metrics are cited from other studies and BloombergGPT is a proprietary model, so the metrics may be obtained from different evaluation settings (e.g., 3/5-classes or different data splits).

The first observation is the different behaviors of GPT-3.5 and BLOOMZ as base models. GPT-3.5 was trained mainly on the Common Crawl corpus~\cite{brown20}, which archives the web. BLOOMZ was trained on an even larger Open-science Open-collaboration Text Sources corpus~\cite{lauren22}, which is mainly crowd-sourced scientific datasets. The five testing datasets are all from the web: this may be closer to GPT-3.5's trained language domain.  It is observed that GPT-3.5 is better instruction-tuned with its proprietary human feedback. In contrast, BLOOMZ inclines to the language completion task. An example is that the prompt ``Translate to English: Je t'aime" without a full stop (.) at the end may result in the model trying to continue the French sentence instead of translating it. BLOOMZ inclines to complete/answer with concise language. For the sentiment-related open-ended questions to heterogeneous agents, BLOOMZ often answers a final judgment of positive/negative without much justification, and is not good at predicting ``neutral" messages. For the afore-discussed factors,

\begin{table}[hbp]
\centering
\caption{Summary statistics of the five FSA datasets (post-processing).}
\begin{tabular}{ l|c|c|c|c|c } 
 \toprule
 \textbf{Dataset} & FPB & StockSen & CMC & FiQA & SEntFiN\\
 \midrule
Positive & 570 & 4542 & 12022 & 507 & 2832  \\ 
Negative & 303 & 1676 & 1523 & 264 & 2373\\ 
Neutral & 1391 & -- & -- & -- & 2701\\ 
 \midrule
Total Size & 2264 & 6218 & 13545 & 771 & 7906\\ 
 \bottomrule
\end{tabular}
\label{stats}
\end{table}

\begin{table}[hbp]
  \caption{Effects of instantiating the HAD design framework on different LLMs.}
  \label{tab:freq}
  \resizebox{\columnwidth}{!}{
  \begin{tabular}{l|cc|cc|cc|cc|cc}
    \toprule
    \multirow{2}{*}{Model\textbackslash{}Dataset} &  \multicolumn{2}{c}{FPB} & \multicolumn{2}{c}{StockSen}& \multicolumn{2}{c}{CMC}& \multicolumn{2}{c}{FiQA}& \multicolumn{2}{c}{SEntFiN}\\
    &Acc.&F-1 &Acc.&F-1&Acc.&F-1&Acc.&F-1&Acc.&F-1\\
    \midrule
    (Fin-)BERT & \textcolor{lightgray}{91.69} & \textcolor{lightgray}{89.70} & \textcolor{lightgray}{76.90} & \textcolor{lightgray}{84.50} & \textcolor{lightgray}{93.50} & -- & -- & -- & \textcolor{lightgray}{94.29} & \textcolor{lightgray}{93.27}\\
    BloombergGPT & -- & \textcolor{lightgray}{51.07} & -- & -- & -- & -- & -- & \textcolor{lightgray}{75.07} & -- & --\\
    GPT-3.5 & 78.58 & 81.06 & 67.64 & 73.93 & 85.31 & 91.05 & 90.53 & 92.41 & 67.99 & 63.21\\
    GPT-3.5 (HAD) & \textbf{80.48} & \textbf{81.41} & \textbf{70.44} & \textbf{76.55} & \textbf{87.55} & \textbf{92.50} & \textbf{93.91} & \textbf{95.22} & \textbf{77.45} & \textbf{76.93}\\
    BLOOMZ & \textbf{34.63} & 32.90 & 63.65 & 72.47 & 87.16 & 92.62 & \textbf{78.33} & \textbf{83.64} & \textbf{51.32} & \textbf{41.87}\\
    BLOOMZ (HAD) & 34.19 & \textbf{32.93} & \textbf{72.80} & \textbf{83.97} & \textbf{87.67} & \textbf{92.95} & 76.78 & 83.03 & 50.16 & 40.69\\
  \bottomrule
\end{tabular}
}
\end{table}
 BLOOMZ performance metrics are generally inferior except for the CMC dataset, and the differences are more pronounced for FPB and SEntFiN, which contain neutral classes.

The second observation is that HAD generally improves the accuracies and F-1 scores on the base models (Table~\ref{tab:freq}). The improvements (from +2.24\% to +9.46\% for accuracy and from +0.35\% to +13.72\% for F1-score) are very consistent on GPT-3.5, probably due to the richer intermediary analysis generated. HAD's effect on BLOOMZ is minimal except for on the StockSen dataset, where the ca. +10\% improvements are significant. Noteworthy, StockSen is the dataset on which the error types for agent design are derived.  

The last observation is on assessing the significance of the improvements. Theoretically, fine-tuning the LLMs to a downstream task will perform better than the ICL/instruction-based/zero-shot setting just as in the differences of supervised/unsupervised learning. The cost of fine-tuning is bi-fold in the context of FSA: you have to ask experts to accumulate and label thousands of examples; and the performance will be fragile to data distribution shifts and dependent on the optimization techniques applied. By comparing the improvements to the overall differences between GPT-3.5 and (Fin)-BERT on FPB, StockSen, CMC, and SEntFiN, a fair estimation is that the HAD framework can fix 25\%--35\% of the gap between ICL and fine-tuning.

\subsection{Ablation Analysis}
To test the importance of each LLM agent, their intermediary responses are removed singly and the performance decreases benchmarked on GPT-3.5 (HAD) are reported in Table~\ref{tab:abl}. Because of time constraints, only three datasets and the average results are used: FPB and FiQA have the two smallest sizes, and the effect of HAD is the most pronounced on SEntFiN.

It is observed that the mood agent (A1), the rhetoric agent (A2), and the aspect agent (A4) are the most important: removing any of them will generally have a negative impact on the performance. The reference agent (A5) is less important: the effect of removing it is uncertain across different datasets. The dependency agent (A3) seems ineffective: removing A3 will further improve the performance. The ineffectiveness of A3 may suggest considering this error type is unnecessary, or be attributed to an ineffective prompt design. Either way, the observed performances suggest that heterogeneous agents have complicated non-linear interactions, and the presented design can be further optimized with more empirical evidence. 

\subsection{Case Study}
Five cases are presented to illustrate the quality of HAD outputs and how those outputs predict a polarity different from naive prompting. \\

\begin{table}[hbp]
\centering
  \caption{Effects of removing one agents (using gpt-3.5-turbo-1106 as the base model)}
  \label{tab:abl}
  \resizebox{\columnwidth}{!}{
  \begin{tabular}{l|cc|cc|cc|cc}
    \toprule
    \multirow{2}{*}{Model\textbackslash{}Dataset} &  \multicolumn{2}{c}{FPB} & \multicolumn{2}{c}{FiQA} &  \multicolumn{2}{c}{SEntFiN} &  \multicolumn{2}{c}{Average}\\
    &Acc.&F-1 &Acc.&F-1 &Acc.&F-1&Acc.&F-1\\
    \midrule
    GPT-3.5 (HAD) & 80.48 & 81.41 & 93.91 & 95.22 & 77.45 & 76.93 & -- & -- \\
    GPT-3.5 & -1.90 & -0.35 & -3.38 & -2.81 & -9.46 & -13.72 & -- & -- \\
    \midrule
    w/o A1 & -0.71 & +0.64 & -0.01 & +0.02 & -0.58 & -0.61 & -0.43 & +0.02 \\
    w/o A2 & -2.12 & -0.39 & +0.64 & +0.52 & -0.80 & -0.99 & -0.76 & -0.29 \\
    w/o A3 & +3.00 & +3.56 & +0.51 & +0.42 & +0.01 & +0.03 & +1.17 & +1.34 \\
    w/o A4 & +0.04 & +0.97 & +0.25 & +0.22  & -0.66 & -0.69 & -0.12 & +0.16 \\
    w/o A5 & +4.32 & +4.29 & -0.01 & -0.00  & -0.52 & -0.43 & +1.26 & +1.28\\
  \bottomrule
\end{tabular}
}
\end{table}

\newpage

In Case 1, multiple companies are mentioned and naive prompting produces a negative prediction without much explanation. With HAD, A1 and A2 believe this message is neutral according to their perspectives. A1's argument is reasonable as the positivity is more directly associated to Wells Fargo than to Berkshire. With A3, A4, A5 considering the message as positive, the framework finally summarizes a correct polarity as positive. 

Case 2 is challenging and can easily be mistaken as positive by naive prompting with key-phrase such as ``drive ... higher" spotted. To correctly understand the context, one has to know that Taylor Wimpey and Ashtead are home construction and construction equipment rental companies. So ``driving the markets higher" may refer to the index or property markets and is setting an economic scenario. It has complicated implications for the two companies and is not as direct as ``Barclays falls". A1, A4, A5 are correct about the mixed sentiment. With A3's judgment being neutral and A2 predicting negative, the framework finally summarizes a correct polarity as negative. 

Case 3 is predicted as positive by naive prompting. Although as A1 explained, less smuggling is good for society, the message is apparently commenting on gold itself as a commodity. Despite the fact that no irrealis mood or any rhetorics are present, A1, A2, A5 correctly predict the message as negative. With A3 and A4 seeing it as neutral, the final decision is negative as per the majority.

\begin{tcolorbox}[colback=white,colframe=black,colbacktitle=black,
title=Case 1: Berkshire applies to boost Wells Fargo stake above 10 pct (positive)]
(\textbf{A1}) Neutral. No indication of positive or negative sentiment. The use of "applies to" suggests an irrealis mood, means Berkshire is seeking to increase its stake but the action has not yet been completed.
(\textbf{A2}) Neutral. No indications of sarcasm or negative assertion in the message.
(\textbf{A3}) The sentiment of the speaker in this message is likely positive, as they are expressing their intention to increase their stake, which suggests confidence.
(\textbf{A4}) The sentiment is positive for "Wells Fargo" as Berkshire Hathaway's application suggests confidence and optimism in the company's potential.
(\textbf{A5}) The sentiment of the message is generally positive. Berkshire Hathaway's decision potentially have a positive impact on Wells Fargo's stock price and overall reputation. The increase timing suggests potential for long-term success despite any current challenges facing Wells Fargo.
\end{tcolorbox}

\begin{tcolorbox}[colback=white,colframe=black,colbacktitle=black,
title={Case 2: London open: Taylor Wimpey and Ashtead drive markets higher, Barclays falls (negative)}]
(\textbf{A1}) Mixed sentiment. Drive markets higher suggests a positive sentiment, while Barclays falls indicates a negative sentiment. No use of irrealis mood.
(\textbf{A2}) Negative. It highlights the fact that Barclays falls, while giving less emphasis to the positive performance of Taylor Wimpey and Ashtead. No specific rhetorics used.
(\textbf{A3}) Neutral. They are simply reporting on the market movements and the performance of certain companies.
(\textbf{A4}) Positive for Taylor Wimpey and Ashtead as they are driving the markets higher, while the sentiment is negative for Barclays as it is falling.
(\textbf{A5}) Overall a mixed sentiment. Some companies are performing well and driving the market higher, others, such as Barclays, are not faring as well.
\end{tcolorbox}

\begin{tcolorbox}[colback=white,colframe=black,colbacktitle=black,
title={Case 3: Smuggling of gold sees a decline as its demand softens (negative)}]
(\textbf{A1}) The sentiment is negative. Word ``decline" indicates a decrease in smuggling activity, which could be seen as positive, but the softening demand for gold implies a negative impact on the overall market for gold. There is no irrealis mood used in this message.
(\textbf{A2}) The sentiment is negative. Decline and soften indicate a decrease in the demand for gold, which could have negative implications for those involved in the smuggling of gold. No specific rhetorical devices present.
(\textbf{A3}) Speaker is neutral.
(\textbf{A4}) Neutral.
(\textbf{A5}) The sentiment is negative. The decrease in the demand indicates a downturn in the market, leading to a decline in smuggling activities. This suggests a weakening gold market, which could impact prices and trade within the industry.
\end{tcolorbox}

\begin{tcolorbox}[colback=white,colframe=black,colbacktitle=black,
title={Case 4: Puravankara's real estate scheme not a CIS: Sebi (negative)}]
(\textbf{A1}) The sentiment is negative. The use of the word ``not" indicates the negation of a positive outcome, and the mention of Sebi suggests that there may be regulatory or legal issues associated with Puravankara's real estate scheme. There does not appear to be any irrealis mood used in this message.
(\textbf{A2}) The sentiment is negative. ``not a CIS" may imply suspicion that Puravankara's real estate scheme was a CIS (Collective Investment Scheme), which is regulated by SEBI (Securities and Exchange Board of India). The use of the term "scheme" also carries a negative connotation, as it can imply a dubious or deceptive plan. Overall, the message communicates a negative view of Puravankara's real estate activities in relation to SEBI regulations.
(\textbf{A3}) Neutral. The statement just delivers factual information.
(\textbf{A4}) Neutral.
(\textbf{A5}) The sentiment appears to be negative, indicated by the mention of Sebi not considering Puravankara's real estate scheme as a Collective Investment Scheme (CIS). There may be some regulatory or compliance issues with the scheme, which could have negative implications for the company and its investors. Additionally, the fact that the scheme is being scrutinized by a regulatory authority implies a potential risk or concern. However, without further details about the specific nature of the scheme or the reasons for Sebi's decision, it is difficult to fully assess the sentiment.
\end{tcolorbox}

\begin{tcolorbox}[colback=white,colframe=black,colbacktitle=black,
title={Case 5: Whirlpool may head to around 450-475: Devang Visaria (neutral)}]
(\textbf{A1}) The sentiment is neutral. The use of "may" indicates an irrealis mood, suggesting that the statement is hypothetical or uncertain.
(\textbf{A2}) The sentiment appears to be neutral. There are no apparent rhetorics or emotional language. The statement simply presents a speculated range for the potential movement of Whirlpool.
(\textbf{A3}) Sentiment of the speaker seems to be somewhat positive as they are mentioning a potential increase in the stock price of Whirlpool, indicating a sense of optimism or anticipation.
(\textbf{A4}) Neutral.
(\textbf{A5}) The sentiment is neutral. It simply provides a prediction regarding the potential price movement without expressing positive or negative. It is based on analysis and does not convey any emotion or bias.
\end{tcolorbox}

Case 4 is difficult to understand and predicted as positive by naive prompting. Jargon and external reference are the main challenges. From the responses of A1, A2, and A5, it can be observed that ``Puravankara is a real estate company", ``CIS means Collective Investment Scheme" and ``Sebi is a security regulatory authority" are shared knowledge. Surprisingly, A2 exhibits temporal and counterfactual reasoning, which is helpful in understanding this message. 

The last Case 5 was wrongly predicted as positive by naive prompting, probably due to the slight positive color of phrasing ``head to". A1 and A2 correctly identified the uncertainty associated with irrealis mood. A3 detected the same positivity as naive prompting, while the other four agents all predict the message as neutral. With the dominant number of neutral predictions (4:1), the framework correctly summarized the polarity as neutral. This shows HAD's capability to correct slight and uncertain sentiments with a discussion mechanism.  

\section{Discussion, Conclusion, and Future Work}
A novel theory-informed LLM collaboration design for FSA, named HAD, is studied. The design involves heterogeneous LLM agents and specializes them with FSA error types discovered in the past literature. This design is more computationally intensive than naive prompting, but has far less complexity compared to many other LLM collaboration frameworks and fine-tuning-based approaches. In view of the research questions and hypotheses, it has been found that HAD effectively improves FSA accuracies across a number of existing datasets, especially when the LLM agents can produce substantial discussions. The design framework fixes around 25\% -- 35\% of the performance gap between prompting and fine-tuning. With error type-based prompts, the LLM agents behave heterogeneously with different focuses. The mood, rhetoric, and aspect agents are more important than the reference agent. The evaluation results support Hypothesis 1 (\textbf{H1}), but reject Hypothesis 2 (\textbf{H2}) with the observation that the performance can be further optimized if the dependency agent is removed.

Technically, this study contributes across two areas in the Knowledge Contribution Framework (KCF) of deep learning in information systems research~\cite{Samtani2023}. The HAD framework is formulated and instantiated in a high-impact application domain of FSA, where fine-tuning is still a dominant paradigm and LLM collaboration is rarely applied. The framework is zero-shot and training-free, therefore, the performance improvement should be able to generalize to other FSA datasets. Practically, financial advisors, traders, fund managers, and other types of investors could use this framework to build their in-house sentiment analysis tools, or extend their knowledge of possible system designs for FSA.  

This preliminary study has a few limitations, which may inspire future research. The first limitation is \emph{scalability}. Predicting or Discussion with LLM agents is slower compared to statistical analysis and incurs costs. For this reason, a large system, i.e., with more agents, is possible, but not preferred during design and evaluation. The second limitation is the \emph{confidentiality} of evaluation datasets. StockSen, CMC, and SEntFiN are relatively new, but FPB and FiQA have been there for quite a few years. Because the training material for LLMs is usually not fully transparent and some LLMs keep updating using reinforcement learning and human feedback, the possibility that the evaluation datasets have been exposed to the LLMs before, causing some information leaks can not be excluded. Finally, the case studies show that the identified error types can almost be solved. It is therefore interesting to explore what are the \emph{reasons} for the new/remaining errors made by LLMs and assess what are the human/expert-level performances on these FSA datasets. 

Some of the unique challenges in FSA, e.g., external references to facts and world knowledge, were thought to be impossible to solve in the short-term future before the transformer architecture models came into existence. With the hope of artificial general intelligence (AGI) around the corner, this study exhibits the versatile capabilities of LLM that are useful for FSA, and calls for more research on this important task.

\bibliographystyle{halpha-abbrv}
\bibliography{neurips_2023}

\begin{thebibliography}{10}
\expandafter\ifx\csname url\endcsname\relax
  \def\url#1{\texttt{#1}}\fi
\expandafter\ifx\csname doi\endcsname\relax
  \def\doi#1{\burlalt{doi:#1}{http://dx.doi.org/#1}}\fi
\expandafter\ifx\csname urlprefix\endcsname\relax\def\urlprefix{URL }\fi
\expandafter\ifx\csname href\endcsname\relax
  \def\href#1#2{#2}\fi
\expandafter\ifx\csname burlalt\endcsname\relax
  \def\burlalt#1#2{\href{#2}{#1}}\fi

\bibitem{Bollen2011}
J.~Bollen, H.~Mao, and X.~Zeng.
\newblock Twitter mood predicts the stock market.
\newblock {\em Journal of Computational Science}, 2(1):1–8, 2011.
\newblock \doi{10.1016/j.jocs.2010.12.007}.

\bibitem{brown20}
T.~B. Brown, B.~Mann, N.~Ryder, M.~Subbiah, J.~Kaplan, P.~Dhariwal,
  A.~Neelakantan, P.~Shyam, G.~Sastry, A.~Askell, S.~Agarwal,
  A.~Herbert{-}Voss, G.~Krueger, T.~Henighan, R.~Child, A.~Ramesh, D.~M.
  Ziegler, J.~Wu, C.~Winter, C.~Hesse, M.~Chen, E.~Sigler, M.~Litwin, S.~Gray,
  B.~Chess, J.~Clark, C.~Berner, S.~McCandlish, A.~Radford, I.~Sutskever, and
  D.~Amodei.
\newblock Language models are few-shot learners.
\newblock In {\em Proceedings of NeuIPS'20}, pages 1877--1901, 2020.
\newblock \doi{10.48550/arXiv.2005.14165}.

\bibitem{chen23}
S.~Chen and F.~Xing.
\newblock Understanding emojis for financial sentiment analysis.
\newblock In {\em Proceedings of ICIS’23}, pages 1--16, 2023.
\newblock
  \urlprefix\url{https://aisel.aisnet.org/icis2023/socmedia_digcollab/socmedia_digcollab/3/}.

\bibitem{chen23b}
X.~Chen, M.~Lin, N.~Schärli, and D.~Zhou.
\newblock Teaching large language models to self-debug, 2023.
\newblock \doi{10.48550/ARXIV.2304.05128}.

\bibitem{chu2022}
L.~Chu, X.-Z. He, K.~Li, and J.~Tu.
\newblock Investor sentiment and paradigm shifts in equity return forecasting.
\newblock {\em Management Science}, 68(6):4301–4325, 2022.
\newblock \doi{10.1287/mnsc.2020.3834}.

\bibitem{cortis2017semeval}
K.~Cortis, A.~Freitas, T.~Daudert, M.~Huerlimann, M.~Zarrouk, S.~Handschuh, and
  B.~Davis.
\newblock Semeval-2017 task 5: Fine-grained sentiment analysis on financial
  microblogs and news.
\newblock In {\em SemEval Workshop}, 2017.
\newblock \doi{10.18653/v1/S17-2089}.

\bibitem{de2018inf}
D.~de~Fran{\c{c}}a~Costa and N.~F.~F. da~Silva.
\newblock Inf-ufg at fiqa 2018 task 1: predicting sentiments and aspects on
  financial tweets and news headlines.
\newblock In {\em Companion Proceedings of the The Web Conference 2018}, pages
  1967--1971, 2018.
\newblock \doi{10.1145/3184558.3191828}.

\bibitem{Deng2023}
J.~Deng, M.~Yang, M.~Pelster, and Y.~Tan.
\newblock Social trading, communication, and networks.
\newblock {\em Information Systems Research}, 2023.
\newblock \doi{10.1287/isre.2021.0143}.

\bibitem{Deng2018}
S.~Deng, Z.~J. Huang, A.~P. Sinha, and H.~Zhao.
\newblock The interaction between microblog sentiment and stock returns: An
  empirical examination.
\newblock {\em MIS Quarterly}, 42(3):895–918, 2018.
\newblock \doi{10.25300/misq/2018/14268}.

\bibitem{deng-www}
X.~Deng, V.~Bashlovkina, F.~Han, S.~Baumgartner, and M.~Bendersky.
\newblock What do llms know about financial markets? a case study on reddit
  market sentiment analysis.
\newblock In {\em Companion Proceedings of the ACM Web Conference 2023}, 2023.
\newblock \doi{10.1145/3543873.3587324}.

\bibitem{Dong2018}
W.~Dong, S.~Liao, and Z.~Zhang.
\newblock Leveraging financial social media data for corporate fraud detection.
\newblock {\em Journal of Management Information Systems}, 35(2):461–487,
  2018.
\newblock \doi{10.1080/07421222.2018.1451954}.

\bibitem{duk2023}
K.~Du, F.~Xing, and E.~Cambria.
\newblock Incorporating multiple knowledge sources for targeted aspect-based
  financial sentiment analysis.
\newblock {\em {ACM} Transactions on Management Information Systems},
  14(3):1--24, 2023.
\newblock \doi{10.1145/3580480}.

\bibitem{md}
Y.~Du, S.~Li, A.~Torralba, J.~B. Tenenbaum, and I.~Mordatch.
\newblock Improving factuality and reasoning in language models through
  multiagent debate, 2023.
\newblock \doi{10.48550/ARXIV.2305.14325}.

\bibitem{fei2023}
H.~Fei, B.~Li, Q.~Liu, L.~Bing, F.~Li, and T.-S. Chua.
\newblock Reasoning implicit sentiment with chain-of-thought prompting.
\newblock In {\em Proceedings of ACL'23}, pages 1171--1182, 2023.
\newblock \doi{10.18653/v1/2023.acl-short.101}.

\bibitem{gregor}
S.~Gregor and A.~R. Hevner.
\newblock Positioning and presenting design science research for maximum
  impact.
\newblock {\em MIS Quarterly}, 37(2):337–355, 2013.
\newblock \doi{10.25300/misq/2013/37.2.01}.

\bibitem{h2021}
T.~Hendershott, X.~M. Zhang, J.~L. Zhao, and Z.~E. Zheng.
\newblock Fintech as a game changer: Overview of research frontiers.
\newblock {\em Information Systems Research}, 32(1):1–17, 2021.
\newblock \doi{10.1287/isre.2021.0997}.

\bibitem{Hendrycks}
D.~Hendrycks, C.~Burns, S.~Basart, A.~Zou, M.~Mazeika, D.~Song, and
  J.~Steinhardt.
\newblock Measuring massive multitask language understanding.
\newblock In {\em Proceedings of ICLR'21}, 2021.
\newblock \doi{10.48550/arXiv.2009.03300}.

\bibitem{hevner}
A.~R. Hevner, S.~T. March, J.~Park, and S.~Ram.
\newblock Design science in information systems research.
\newblock {\em MIS Quarterly}, 28(1):75--105, 2004.
\newblock \doi{10.2307/25148625}.

\bibitem{lauren22}
H.~Lauren{\c{c}}on, L.~Saulnier, T.~Wang, C.~Akiki, A.~V. del Moral, T.~L.
  Scao, L.~von Werra, C.~Mou, E.~G. Ponferrada, H.~Nguyen, J.~Frohberg,
  M.~Sasko, Q.~Lhoest, A.~McMillan{-}Major, G.~Dupont, S.~Biderman, A.~Rogers,
  L.~B. Allal, F.~D. Toni, G.~Pistilli, O.~Nguyen, S.~Nikpoor, M.~Masoud,
  P.~Colombo, J.~de~la Rosa, P.~Villegas, T.~Thrush, S.~Longpre, S.~Nagel,
  L.~Weber, M.~Mu{\~{n}}oz, J.~Zhu, D.~van Strien, Z.~Alyafeai, K.~Almubarak,
  M.~C. Vu, I.~Gonzalez{-}Dios, A.~Soroa, K.~Lo, M.~Dey, P.~O. Suarez,
  A.~Gokaslan, S.~Bose, D.~I. Adelani, L.~Phan, H.~Tran, I.~Yu, S.~Pai,
  J.~Chim, V.~Lepercq, S.~Ilic, M.~Mitchell, A.~S. Luccioni, and Y.~Jernite.
\newblock The bigscience {ROOTS} corpus: {A} 1.6tb composite multilingual
  dataset.
\newblock In {\em Proceedings of NeurIPS'22}, 2022.
\newblock \doi{10.48550/arXiv.2303.03915}.

\bibitem{lengkeek}
M.~Lengkeek, F.~{van der Knaap}, and F.~Frasincar.
\newblock Leveraging hierarchical language models for aspect-based sentiment
  analysis on financial data.
\newblock {\em Information Processing \& Management}, 60(5):103435, 2023.
\newblock \doi{https://doi.org/10.1016/j.ipm.2023.103435}.

\bibitem{liupf}
P.~Liu, W.~Yuan, J.~Fu, Z.~Jiang, H.~Hayashi, and G.~Neubig.
\newblock Pre-train, prompt, and predict: A systematic survey of prompting
  methods in natural language processing.
\newblock {\em {ACM} Computing Surveys}, 55(9):1–35, 2023.
\newblock \doi{10.1145/3560815}.

\bibitem{Liu2022}
V.~Liu and L.~B. Chilton.
\newblock Design guidelines for prompt engineering text-to-image generative
  models.
\newblock In {\em Proceedings of CHI ’22}, 2022.
\newblock \doi{10.1145/3491102.3501825}.

\bibitem{Liu2020}
Z.~Liu, D.~Huang, K.~Huang, Z.~Li, and J.~Zhao.
\newblock Finbert: A pre-trained financial language representation model for
  financial text mining.
\newblock In {\em Proceedings of IJCAI'20}, pages 4513--4519, 2020.
\newblock \doi{10.24963/ijcai.2020/622}.

\bibitem{loughran2011}
T.~Loughran and B.~McDonald.
\newblock When is a liability not a liability? textual analysis, dictionaries,
  and 10-ks.
\newblock {\em Journal of Finance}, 66(1):35--65, 2011.
\newblock \doi{10.1111/j.1540-6261.2010.01625.x}.

\bibitem{maia201818}
M.~Maia, S.~Handschuh, A.~Freitas, B.~Davis, R.~McDermott, M.~Zarrouk, and
  A.~Balahur.
\newblock {WWW}'18 open challenge: financial opinion mining and question
  answering.
\newblock In {\em Proceedings of WWW'18}, pages 1941--1942, 2018.
\newblock \doi{10.1145/3184558.3192301}.

\bibitem{malo2014good}
P.~Malo, A.~Sinha, P.~Korhonen, J.~Wallenius, and P.~Takala.
\newblock Good debt or bad debt: Detecting semantic orientations in economic
  texts.
\newblock {\em Journal of the Association for Information Science and
  Technology}, 65(4):782--796, 2014.
\newblock \doi{10.1002/asi.23062}.

\bibitem{emotion-machine}
M.~Minsky.
\newblock {\em The Emotion Machine: Commonsense Thinking, Artificial
  Intelligence, and the Future of the Human Mind}.
\newblock Simon \& Schuster, 2006.

\bibitem{bloomz}
N.~Muennighoff, T.~Wang, L.~Sutawika, A.~Roberts, S.~Biderman, T.~Le~Scao,
  M.~S. Bari, S.~Shen, Z.~X. Yong, H.~Schoelkopf, X.~Tang, D.~Radev, A.~F. Aji,
  K.~Almubarak, S.~Albanie, Z.~Alyafeai, A.~Webson, E.~Raff, and C.~Raffel.
\newblock Crosslingual generalization through multitask finetuning.
\newblock In {\em Proceedings of ACL'23}, pages 15991--16111, 2023.
\newblock \doi{10.18653/v1/2023.acl-long.891}.

\bibitem{nori23}
H.~Nori, Y.~T. Lee, S.~Zhang, D.~Carignan, R.~Edgar, N.~Fusi, N.~King,
  J.~Larson, Y.~Li, W.~Liu, R.~Luo, S.~M. McKinney, R.~O. Ness, H.~Poon,
  T.~Qin, N.~Usuyama, C.~White, and E.~Horvitz.
\newblock Can generalist foundation models outcompete special-purpose tuning?
  case study in medicine, 2023.
\newblock \doi{10.48550/ARXIV.2311.16452}.

\bibitem{Peterson2016}
R.~L. Peterson.
\newblock {\em Trading on Sentiment: The Power of Minds Over Markets}.
\newblock Wiley, 2016.
\newblock \doi{10.1002/9781119219149}.

\bibitem{Samtani2023}
S.~Samtani, H.~Zhu, B.~Padmanabhan, Y.~Chai, H.~Chen, and J.~F. Nunamaker.
\newblock Deep learning for information systems research.
\newblock {\em Journal of Management Information Systems}, 40(1):271–301,
  2023.
\newblock \doi{10.1080/07421222.2023.2172772}.

\bibitem{shah2022}
R.~Shah, K.~Chawla, D.~Eidnani, A.~Shah, W.~Du, S.~Chava, N.~Raman, C.~Smiley,
  J.~Chen, and D.~Yang.
\newblock When {FLUE} meets {FLANG}: Benchmarks and large pretrained language
  model for financial domain.
\newblock In {\em Proceedings of EMNLP'22}, 2022.
\newblock \doi{10.18653/v1/2022.emnlp-main.148}.

\bibitem{sentfin}
A.~Sinha, S.~Kedas, R.~Kumar, and P.~Malo.
\newblock {SE}nt{F}i{N} 1.0: {E}ntity‐aware sentiment analysis for financial
  news.
\newblock {\em Journal of the Association for Information Science and
  Technology}, 73(9):1314–1335, 2022.
\newblock \doi{10.1002/asi.24634}.

\bibitem{Sorensen2022}
T.~Sorensen, J.~Robinson, C.~Rytting, A.~Shaw, K.~Rogers, A.~Delorey,
  M.~Khalil, N.~Fulda, and D.~Wingate.
\newblock An information-theoretic approach to prompt engineering without
  ground truth labels.
\newblock In {\em Proceedings of ACL'22}, 2022.
\newblock \doi{10.18653/v1/2022.acl-long.60}.

\bibitem{mn}
X.~Sun, X.~Li, S.~Zhang, S.~Wang, F.~Wu, J.~Li, T.~Zhang, and G.~Wang.
\newblock Sentiment analysis through llm negotiations, 2023.
\newblock \doi{10.48550/ARXIV.2311.01876}.

\bibitem{bloomberggpt}
S.~Wu, O.~Irsoy, S.~Lu, V.~Dabravolski, M.~Dredze, S.~Gehrmann, P.~Kambadur,
  D.~Rosenberg, and G.~Mann.
\newblock Bloomberggpt: A large language model for finance, 2023.
\newblock \doi{10.48550/ARXIV.2303.17564}.

\bibitem{xing2018}
F.~Xing, E.~Cambria, and R.~Welsch.
\newblock Intelligent asset allocation via market sentiment views.
\newblock {\em IEEE Computational Intelligence Magazine}, 13(4):25–34, 2018.
\newblock \doi{10.1109/mci.2018.2866727}.

\bibitem{saving}
F.~Xing, E.~Cambria, and Y.~Zhang.
\newblock Sentiment-aware volatility forecasting.
\newblock {\em Knowledge Based Systems}, 176:68--76, 2019.
\newblock \doi{10.1016/j.knosys.2019.03.029}.

\bibitem{fsa}
F.~Xing, L.~Malandri, Y.~Zhang, and E.~Cambria.
\newblock Financial sentiment analysis: An investigation into common mistakes
  and silver bullets.
\newblock In {\em Proceedings of COLING’20}, pages 978--987, 2020.
\newblock \doi{10.18653/v1/2020.coling-main.85}.

\bibitem{xing2019cognitive}
F.~Xing, F.~Pallucchini, and E.~Cambria.
\newblock Cognitive-inspired domain adaptation of sentiment lexicons.
\newblock {\em Information Processing \& Management}, 56(3):554--564, 2019.
\newblock \doi{10.1016/j.ipm.2018.11.002}.

\bibitem{yang2023}
Y.~Yang, Y.~Qin, Y.~Fan, and Z.~Zhang.
\newblock Unlocking the power of voice for financial risk prediction: A
  theory-driven deep learning design approach.
\newblock {\em MIS Quarterly}, 47(1):63–96, 2023.
\newblock \doi{10.25300/misq/2022/17062}.

\bibitem{tot}
S.~Yao, D.~Yu, J.~Zhao, I.~Shafran, T.~Griffiths, Y.~Cao, and K.~Narasimhan.
\newblock Tree of thoughts: Deliberate problem solving with large language
  models.
\newblock In {\em Proceedings of NeuIPS’23}, pages 1--14, 2023.

\bibitem{yu2023}
F.~Yu, L.~Quartey, and F.~Schilder.
\newblock Exploring the effectiveness of prompt engineering for legal reasoning
  tasks.
\newblock In {\em Findings of the Association for Computational Linguistics},
  2023.
\newblock \doi{10.18653/v1/2023.findings-acl.858}.

\bibitem{zhang-icaif}
B.~Zhang, H.~Yang, T.~Zhou, M.~Ali~Babar, and X.-Y. Liu.
\newblock Enhancing financial sentiment analysis via retrieval augmented large
  language models.
\newblock In {\em Proceedings of ICAIF’23}, 2023.
\newblock \doi{10.1145/3604237.3626866}.

\bibitem{zhangd}
W.~Zhang, Y.~Deng, B.~Liu, S.~J. Pan, and L.~Bing.
\newblock Sentiment analysis in the era of large language models: A reality
  check, 2023.
\newblock \doi{10.48550/ARXIV.2305.15005}.

\bibitem{Zimbra}
D.~Zimbra, A.~Abbasi, D.~Zeng, and H.~Chen.
\newblock The state-of-the-art in twitter sentiment analysis: {A} review and
  benchmark evaluation.
\newblock {\em {ACM} Transactions on Management Information Systems},
  9(2):5:1--5:29, 2018.
\newblock \doi{10.1145/3185045}.

\end{thebibliography}

\end{document}